\documentclass[10pt,twocolumn,letterpaper]{article}

\usepackage{iccv}
\usepackage{epsfig}
\usepackage{graphicx}
\usepackage{amsmath}
\usepackage{amssymb}
\usepackage{subfigure}
\usepackage{multirow}
\usepackage[labelfont={bf, rm},font={small,it}]{caption}

\def\T{{\!\top}}

\iccvfinalcopy %

\ificcvfinal\fi

\begin{document}

\title{Towards End-to-end Text Spotting  with Convolutional Recurrent Neural Networks}

\author{Hui Li, Peng Wang, Chunhua Shen\thanks{Corresponding author, e-mail:\tt chhshen@gmail.com }\\
Machine Learning Group, The University of Adelaide, Australia}

\maketitle
\begin{abstract}
In this work, we jointly address the problem of text detection and recognition in natural scene images based on convolutional recurrent neural networks.
We propose a unified network that simultaneously localizes and recognizes text with a single forward pass, avoiding intermediate processes like image cropping and feature re-calculation, word separation, or character grouping.
In contrast to existing approaches that consider text detection and recognition as two distinct tasks and tackle them one by one, the proposed framework settles these two tasks concurrently.
The whole framework can be trained end-to-end, requiring only images, the ground-truth bounding boxes and text labels.  Through end-to-end training, the learned features can be more informative, which improves the overall performance. The convolutional features are calculated only once and shared by both detection and recognition, which saves processing time.
Our proposed method has achieved competitive performance on several benchmark datasets.

\end{abstract}

\tableofcontents
\clearpage

\section{Introduction}
\label{sec:intro}

Text in natural scene images contains rich semantic information and is of great value for image understanding.
As an important task in image analysis, scene text spotting, including both text detection and word recognition, attracts much attention in computer vision field. It has many potential applications, ranging from web image searching, robot navigation, to image retrieval.

Due to the large variability of text patterns and the highly complicated background, text spotting in natural scene images is much more challenging than from scanned documents. Although significant progress has been made recently based on Deep Neural Network (DNN) techniques,
it is still an open problem~\cite{xiang2016Survey}.

Previous works~\cite{Wangkai2011, Bissacco2013ICCV, Max2014ECCV, Max2016IJCV} usually
divide text spotting into a sequence of distinct sub-tasks. Text detection is performed firstly with a high recall to get candidate regions of text. Then word recognition is applied on the cropped text bounding boxes by a totally different approach,  following word separation or character grouping.
A number of techniques are also developed which solely focus on text detection or word recognition.
However, the tasks of word detection and recognition are highly correlated. Firstly, the feature information can be shared between them. In addition, these two tasks can complement each other:
better detection improves recognition accuracy,
and the recognition information can refine detection results vice versa.

\begin{figure*}
	\begin{center}
		\includegraphics[width=0.9\textwidth]{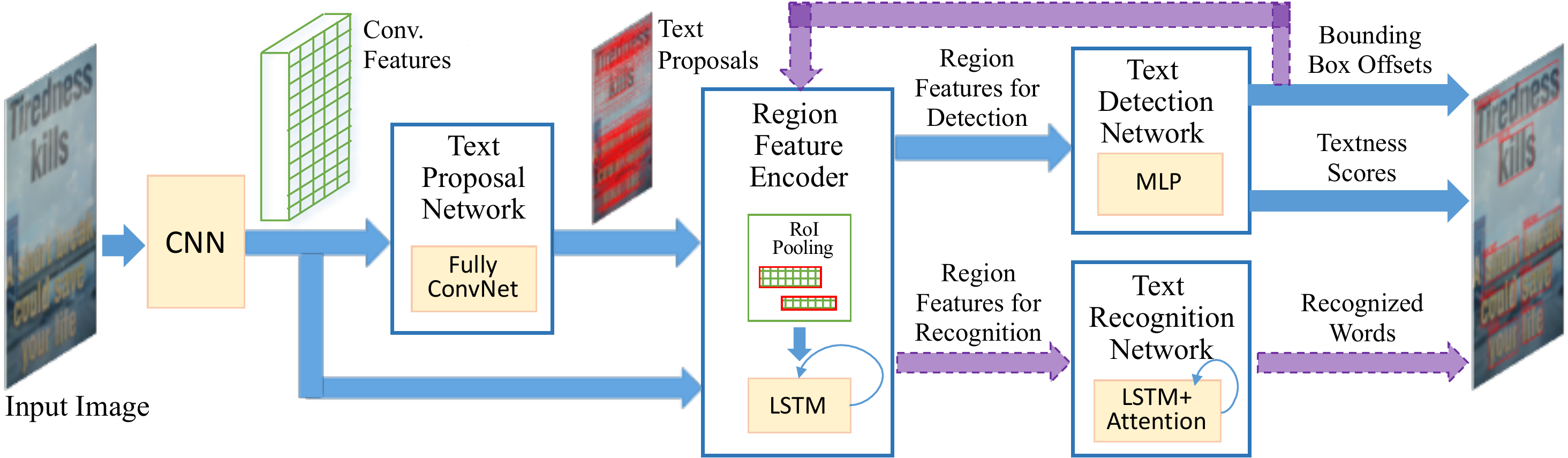}
	\end{center}
	\caption{Model overview. The network takes an image as input, and outputs both text bounding boxes and text labels in one forward pass. The whole network is trained end-to-end.}
	\label{fig:overview}
\end{figure*}

To this end, we propose an end-to-end trainable text spotter, which jointly detects and recognizes words in an image. An overview of the network architecture is presented in Figure~\ref{fig:overview}. It consists of
a number of convolutional layers, a region proposal network tailored specifically for text (refer to as Text Proposal Network, TPN), an Recurrent Neural Network (RNN) encoder for embedding proposals of varying sizes to fixed-length vectors, multi-layer perceptrons for detection and bounding box regression,
and an attention-based RNN decoder for word recognition.
Via this framework, both text bounding boxes and word labels are provided with a single forward evaluation of the network.  We do not need to process the intermediate issues such as character grouping~\cite{Zhu_2016_CVPR, Textflow2015ICCV} or text line separation~\cite{zhengCVPR15}, and thus avoid error accumulation.
The main contributions are thus three-fold.

 $(1)$ An end-to-end trainable DNN is designed to optimize the overall accuracy and share computations. The network integrates both text detection and word recognition. With the end-to-end training of both tasks, the learned features are more informative, which can promote the detection performance as well as the overall performance. The convolutional features are shared by both detection and recognition, which saves processing time.
To our best knowledge, this is the first attempt to integrate text detection and recognition into a single end-to-end trainable network.

 $(2)$ We propose a new method for region feature extraction. In previous works~\cite{RGB2015ICCV, renNIPS15fasterrcnn}, Region-of-Interest (RoI) pooling layer converts regions of different sizes and aspect ratios into feature maps with a fixed size.
Considering the significant diversity of aspect ratios in text bounding boxes, it is sub-optimal to fix the size after pooling.
To accommodate the original aspect ratios and avoid distortion, RoI pooling is tailored to generate feature maps with varying lengths. An RNN encoder is then
employed to encode feature maps of different lengths into the same size.

 $(3)$ A curriculum learning strategy is designed to train the system
with gradually more complex training data.
Starting from synthetic images with simple appearance and a large word lexicon, the system learns a character-level language model
and finds a good initialization of appearance model.
By employing real-world images with a small lexicon later, the system gradually
learns how to handle complex appearance patterns.
we conduct a set of experiments to explore the capabilities of different model structures. The best model outperforms state-of-the-art results on a number of standard text spotting benchmarks, including ICDAR2011, 2015.

\section{Related Work}
\label{sec:ReWork}

Text spotting essentially includes two tasks: text detection and word recognition. In this section, we present a brief introduction to related works on text detection, word recognition, and text spotting systems that combine both. There are comprehensive surveys for text detection and recognition in~\cite{Ye2015pami,xiang2016Survey}.

\subsection{Text Detection} Text detection aims to localize text in images and generate bounding boxes for words. Existing approaches can be roughly classified into three categories: character based, text-line based and word based methods.

Character based methods firstly find characters in images, and then group them into words.
They can be further divided into sliding window based~\cite{Max2014ECCV,Wang2012, Zhu_2016_CVPR, Textflow2015ICCV} and Connected Components (CC) based~\cite{huang2013ICCV,Neumann2013ICCV, Busta_2015_ICCV} methods. Sliding window based approaches use a trained classifier to detect characters across the image in a multi-scale sliding window fashion. CC based methods segment pixels with consistent region properties (\ie, color, stroke width, density, \etc) into characters. The detected characters are further grouped into text regions by morphological operations, CRF or other graph models.

Text-line based methods detect text lines firstly and then separate each line into multiple words. The motivation is that people usually distinguish text regions initially even if characters are not recognized.
Based on the observation that a text region usually exhibits high self-similarity to itself and strong contrast to its local background,
Zhang~\etal~\cite{zhengCVPR15} propose to extract text lines by exploiting symmetry property.
Zhang~\etal~\cite{Zhang_2016_CVPR} localize text lines via salient maps that are calculated by fully convolutional networks. Post-processing techniques are also proposed in~\cite{Zhang_2016_CVPR} to extract text lines in multiple orientations.

More recently, a number of approaches are proposed to detect words directly in the images using DNN based techniques, such as Faster R-CNN~\cite{renNIPS15fasterrcnn}, YOLO~\cite{YOLO2016}, SSD~\cite{SSD2016}.
By extending Faster R-CNN, Zhong~\etal~\cite{Zhong2016} design a text detector with a multi-scale Region Proposal Network (RPN) and a multi-level RoI pooling layer. Tian~\etal~\cite{Tian2016} develop a vertical anchor mechanism, and propose a Connectionist Text Proposal Network (CTPN) to accurately localize text lines in natural image.
Gupta~\etal~\cite{Gupta16} use a Fully-Convolutional Regression Network (FCRN) for efficient text detection and bounding box regression, motivated by YOLO.
Similar to SSD, Liao~\etal~\cite{LiaoAAAi2017} propose ``TextBoxes''  by combining predictions from multiple feature maps with different resolutions, and achieve the best-reported performance on text detection with datasets in~\cite{icdar2015, Wangkai2011}.

\subsection{Text Recognition}

Traditional approaches to text recognition usually perform in a bottom-up fashion, which recognize individual characters firstly and then integrate them into words by means of beam
search~\cite{Bissacco2013ICCV}, dynamic programming~\cite{Max2014ECCV}, \etc. In contrast, Jaderberg~\etal~\cite{maxNIPS14} consider word recognition as a multi-class classification problem,
and categorize each word over a large dictionary (about $90$K words, \ie, class labels) using a deep CNN.

With the success of RNNs on handwriting recognition~\cite{Graves2006ICML}, He~\etal~\cite{He2015Reading} and Shi~\etal~\cite{ShiBY15} solve word recognition as a sequence labelling problem. RNNs are employed to generate sequential labels of arbitrary length without character segmentation, and Connectionist Temporal Classification (CTC) is adopted to decode the sequence.
Lee and Osindero~\cite{Lee_2016_CVPR} and Shi~\etal~\cite{shiCVPR2016} propose to recognize text using an attention-based sequence-to-sequence learning structure. In this manner, RNNs automatically learn the character-level language model presented in word strings from the training data. The soft-attention mechanism allows the model to selectively exploit local image features. These networks can be trained end-to-end with cropped word image patches as input. Moreover, Shi~\etal~\cite{shiCVPR2016} insert a Spatial Transformer Network (STN) to handle words with irregular shapes.

\subsection{Text Spotting Systems}

Text spotting needs to handle both text detection and word recognition.
Wang~\etal~\cite{Wangkai2011} take the locations and scores of detected characters as input and try to find an optimal configuration of a particular word in a given lexicon, based on a pictorial structures formulation. Neumann and Matas~\cite{Neumann2013ICCV} use a CC based method for character detection. These characters are then agglomerated into text lines based on heuristic rules. Optimal sequences are finally found in each text line using dynamic programming, which are the recognized words. These recognition-based pipelines lack explicit word detection.

Some text spotting systems firstly generate text proposals with a high recall and a low precision, and then refine them during recognition with a separate model. It is expected that a strong recognizer can reject false positives, especially when a lexicon is given. Jaderberg~\etal~\cite{Max2016IJCV} use an ensemble model to generate text proposals, and then adopt the word classifier in~\cite{maxNIPS14} for recognition. Gupta~\etal~\cite{Gupta16} employ FCRN for text detection and the word classifier in~\cite{maxNIPS14} for recognition.
Most recently, Liao~\etal~\cite{LiaoAAAi2017} combine ``TextBoxes'' and ``CRNN''~\cite{ShiBY15}, which yield state-of-the-art performance on text spotting task with datasets in~\cite{icdar2015, Wangkai2011}.

\section{Model}
\label{sec:Model}

Our goal is to design an end-to-end trainable network, which simultaneously detects and recognizes all words in images. Our model is motivated by recent progresses in DNN models such as Faster R-CNN~\cite{renNIPS15fasterrcnn} and sequence-to-sequence learning~\cite{shiCVPR2016, Lee_2016_CVPR}, but we take the special characteristics of text into consideration. In this section, we present a detailed description of the whole system.

\noindent {\bf Notation} All bold capital letters represent matrices and
all bold lower-case letters denote column vectors. $[\mathbf{a};\mathbf{b}]$ concatenates the vectors $\mathbf{a}$ and
$\mathbf{b}$ vertically, while $[\mathbf{a},\mathbf{b}]$ stacks $\mathbf{a}$ and $\mathbf{b}$ horizontally (column wise). In the following, the bias terms in neural networks are omitted.

\subsection{Overall Architecture}
The whole system architecture is illustrated in Figure~\ref{fig:overview}.
Firstly, the input image is fed into a convolutional neural network that is  modified from VGG-$16$ net~\cite{Simonyan14c}.
VGG-$16$ consists of $13$ layers of $3 \times 3$ convolutions followed by Rectified Linear Unit (ReLU),
$5$ layers of $2 \times 2 $ max-pooling, and Fully-Connected (FC) layers.
Here we remove FC layers. As long as text in images can be relatively small, we only keep the $1$st, $2$nd and $4$th layers of max-pooling,
so that the down-sampling ratio is increased from $1/32$ to $1/8$.

Given the computed convolutional features, TPN provides a list of text region proposals (bounding boxes).
Then, Region Feature Encoder (RFE) converts the convolutional
features of proposals into fixed-length representations.
These representations are further fed into Text Detection Network (TDN) to calculate their textness scores and bounding box offsets.
Next, RFE is applied again to compute fixed-length representations of text bounding boxes provided by TDN (see purple paths in Figure~\ref{fig:overview}).
Finally, Text Recognition Network (TRN) recognizes
words in the detected bounding boxes based on their representations.

\subsection{Text Proposal Network}

\begin{figure}[t!]
	\vspace{-0.0cm}
	\centering
	\begin{minipage}[t]{0.26\textwidth}
		\centering
		\includegraphics[width=0.82\textwidth]{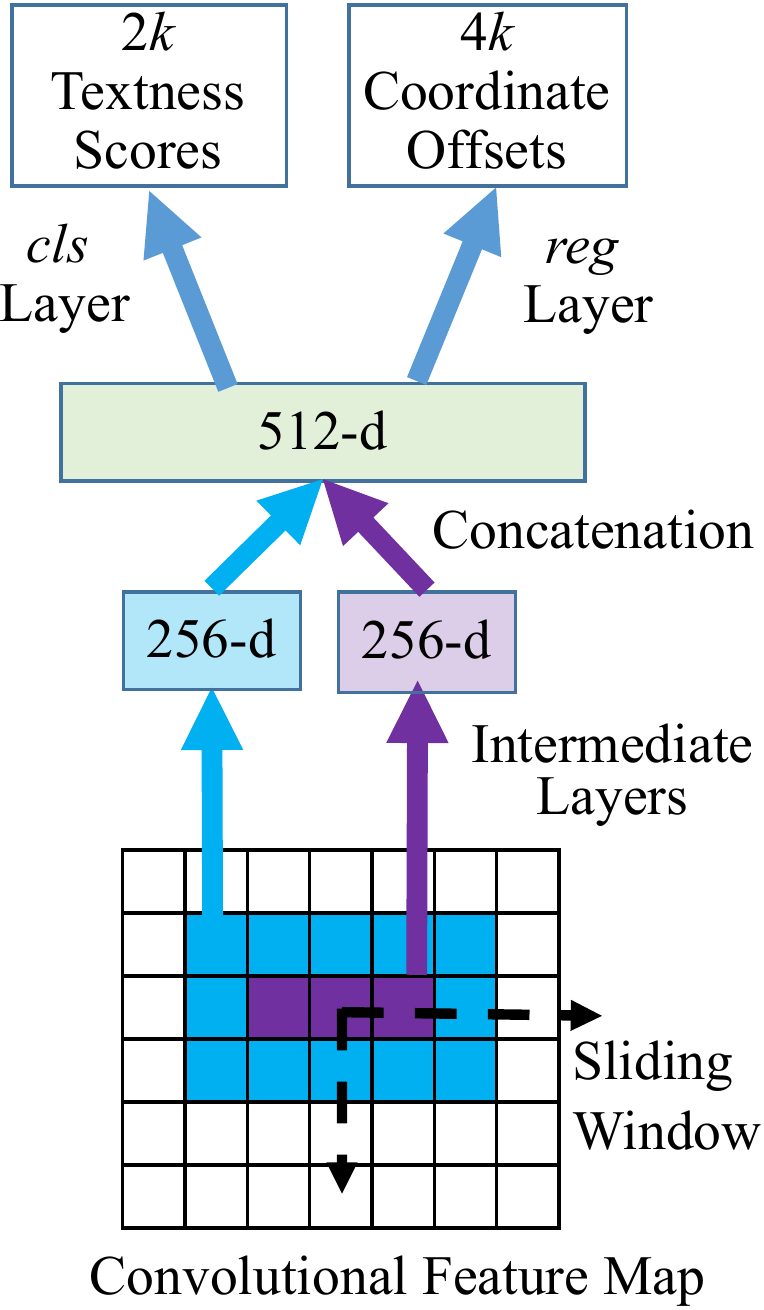}
\end{minipage}\hfill
\begin{minipage}[t]{0.21\textwidth}
	\centering
	\vspace{-6.4cm}
	\caption{Text Proposal Network (TPN). We apply multiple scale sliding windows over the convolutional feature maps. Both local and contextual information are retained which helps to propose high quality text bounding boxes. The concatenated local and contextual features are further fed into the {\em cls} layer for computing textness scores and the {\em reg} layer
	to calculate coordinate offsets, with respect to $k$ anchors at each position.}
	\label{fig:TPN}
\end{minipage}
\end{figure}

Text proposal network (TPN) is inspired from RPN~\cite{renNIPS15fasterrcnn, Zhong2016}, which can be regarded as a fully convolutional network. As presented in Figures~\ref{fig:TPN}, it
takes convolutional features as input, and outputs a set of bounding boxes accompanied with ``textness'' scores and coordinate offsets which indicate scale-invariant translations and log-space height/width shifts relative to pre-defined anchors, as in \cite{renNIPS15fasterrcnn}.

Considering that word bounding boxes usually have larger aspect ratios ($W/H$) and varying scales, we designed $k=24$ anchors with $4$ scales (with box areas of $16^2$, $32^2$, $64^2$, $80^2$) and $6$ aspect ratios ($1:1$, $2:1$, $3:1$, $5:1$, $7:1$, $10:1$).

Inspired by~\cite{Zhong2016}, we apply two $256$-d rectangle convolutional filters of different sizes ($W=5,H=3$ and $W=3,H=1$) on the feature maps to extract both local and contextual information. The rectangle filters lead to wider receptive fields, which is more suitable for word bounding boxes with large aspect ratios. The resulting features are further concatenated to $512$-d vectors and fed into two sibling layers for text/non-text classification and bounding box regression.

\subsection{Region Feature Encoder }
\label{sec:rfe}

\begin{figure}[t]
	\begin{center}
		\includegraphics[width=0.47\textwidth]{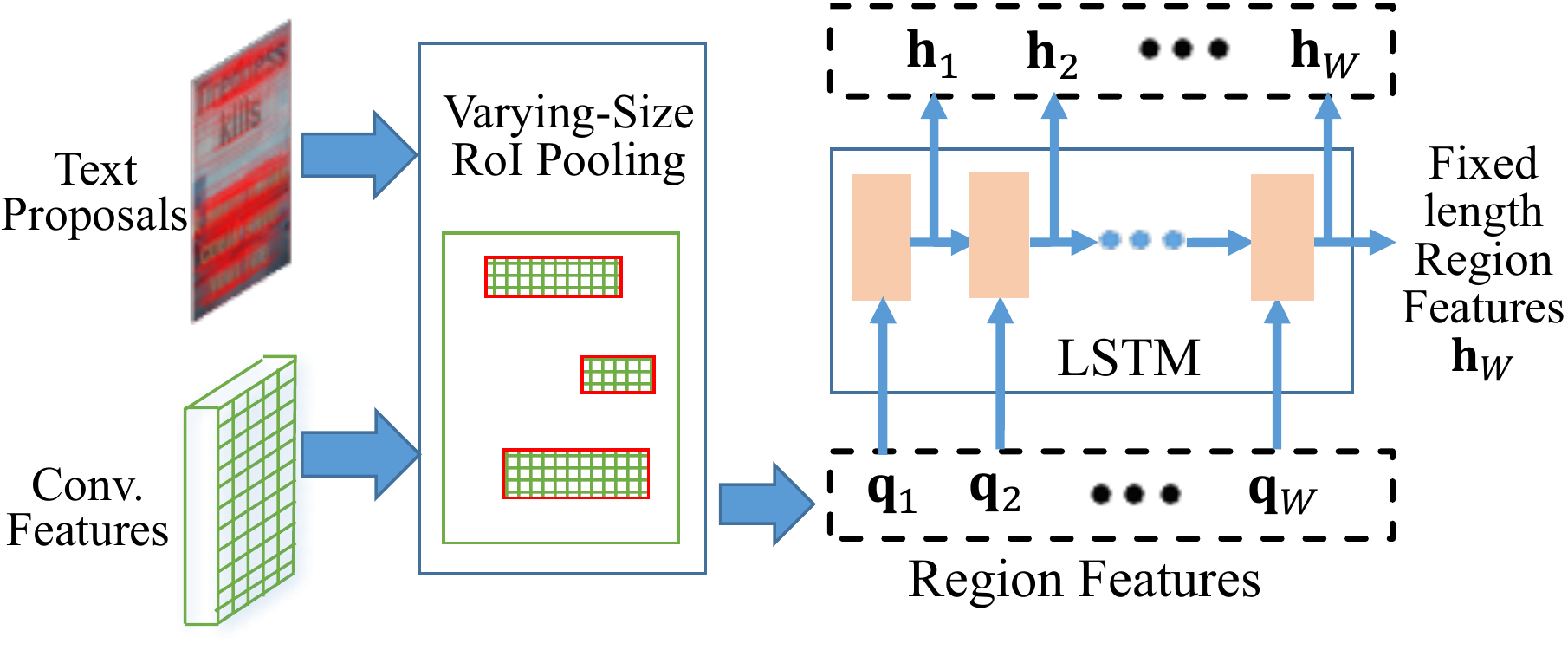}
	\vspace{-0.2cm}
	\end{center}
	\caption{Region Features Encoder (RFE). The region features after RoI pooling are not required to be of the same size. In contrast, they are calculated according to aspect ratio of each bounding box, with height normalized. LSTM is then employed to encode different length region features into the same size.}
	\label{fig:ROI}
		\vspace{-0.3cm}
\end{figure}

To process RoIs of different scales and aspect ratios in a unified way,
most existing works re-sample regions into {\em fixed-size} feature maps via pooling~\cite{RGB2015ICCV}.
However, for text, this approach may lead to significant distortion due to the large variation of word lengths. For example, it may be unreasonable to encode short words like ``Dr'' and long words like ``congratulations'' into feature maps of the same size.
In this work, we propose to re-sample regions according to their respective aspect ratios, and then use RNNs to encode the resulting feature maps of different lengths into fixed length vectors.
The whole region feature encoding process is illustrated in Figure~\ref{fig:ROI}.

For an RoI of size $h \times w$, we perform spatial max-pooling with a resulting size of
\begin{equation}
H \times \min(W_{max},2Hw/h), \label{eq:pool}
\end{equation}
where the expected height $H$ is fixed and the width is adjusted to keep
the aspect ratio as $2w/h$ (twice the original aspect ratio) unless it exceeds the maximum length $W_{max}$.  Note that here we employ a pooling window with an aspect ratio of $1:2$, which benefits the recognition of narrow shaped characters, like `i', `l', etc., as stated in~\cite{ShiBY15}.

Next, the resampled feature maps are considered as a sequence
and fed into RNNs for encoding.
 Here we use Long-Short Term Memory (LSTM)~\cite{LSTM} instead of vanilla RNN to overcome the shortcoming of gradient vanishing or exploding.
The feature maps after the above varying-size RoI pooling are denoted as $\mathbf{Q} \in \mathbb{R}^{C \times H \times W}$, where $W=\min(W_{max},2Hw/h)$ is the number of columns and $C$ is the channel size.
We flatten the features in each column, and obtain a sequence $\mathbf{q}_1, \dots, \mathbf{q}_W \in \mathbb{R}^{C \times H}$ which are fed into LSTMs one by one. Each time LSTM units receive one column of feature $\mathbf{q}_t$, and update their hidden state $\mathbf{h}_t$ by a non-linear function: $\mathbf{h}_t =\mathrm{f} (\mathbf{q}_t, \mathbf{h}_{t-1})$. In this recurrent fashion, the final hidden state $\mathbf{h}_W$ (with size $R = 1024$) captures the holistic information of $\mathbf{Q}$
and is used as a RoI representation with fixed dimension.

\subsection{Text Detection and Recognition}
\label{sec:trn}

\noindent {\bf Text Detection Network} (TDN) aims to judge whether the proposed RoIs are text or not and refine the coordinates of bounding boxes once again, based on the extracted region features $\mathbf{h}_W$.
Two fully-connected layers with $2048$ neurons
are applied on $\mathbf{h}_W$,
followed by two parallel layers for
classification and bounding box regression respectively.

The classification and regression layers used in TDN are similar to those
used in TPN.
Note that the whole system refines the
coordinates of text bounding boxes twice: once in TPN and
then in TDN.
Although RFE is employed twice to calculate features for proposals produced
by TPN and later the detected bounding boxes provided by TDN,
the convolutional features only need to be computed once.

\begin{figure}[t]
	\begin{center}
		\includegraphics[width=0.47\textwidth]{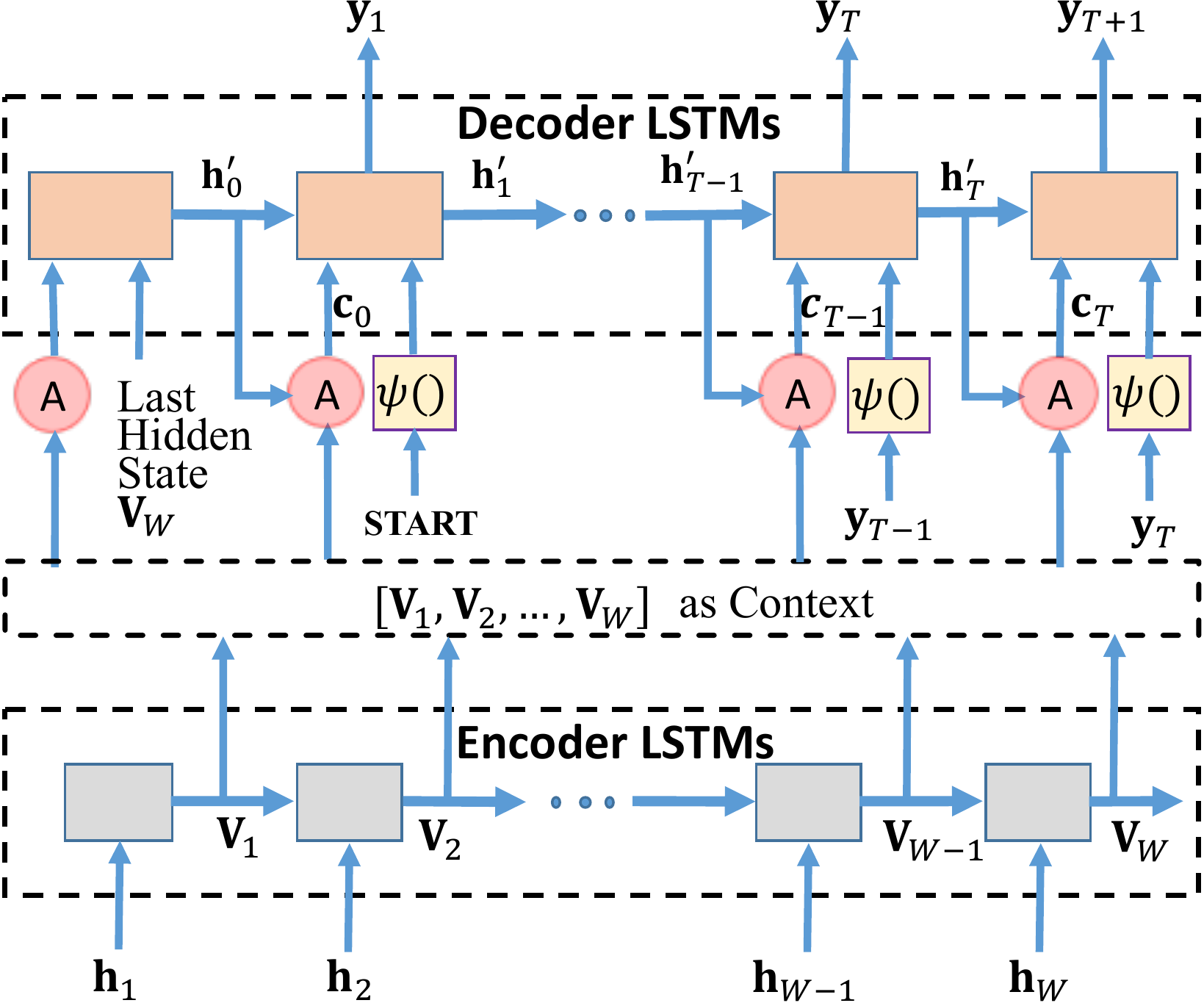}
	\end{center}
	\caption{Text Recognition Network (TRN). The region features are encoded by one layer of LSTMs, and then decoded in an attention based sequence to sequence manner.  Hidden states of encoder at all time steps are reserved and used as context for attention model.
	}
	\label{fig:recognition}
\end{figure}

\noindent {\bf Text Recognition Network} (TRN) aims to predict the text in the detected bounding boxes based on the extracted region features.
As shown in Figure~\ref{fig:recognition},
we adopt LSTMs with attention mechanism~\cite{luong2015EMNLP, shiCVPR2016}
to decode the sequential features into words.

Firstly, hidden states at all steps $\mathbf{h}_1, \dots, \mathbf{h}_W$ from RFE are fed into an additional layer of LSTM encoder with $1024$ units.
We record the hidden state at each time step and form a sequence of $\mathbf{V} =[\mathbf{v}_1, \dots, \mathbf{v}_W]\in \mathbb{R}^{R \times W}$. It includes local information at each time step and works as the context for the attention model.

As for decoder LSTMs, the ground-truth word label is adopted as input during training. It can be regarded as a sequence of tokens $\mathbf{s} = \{s_0, s_1, \dots, s_{T+1} \}$ where $s_0$ and $s_{T+1}$ represent the special tokens START and END respectively. We feed decoder LSTMs with $T+2$ vectors:
$\mathbf{x}_{0}$, $\mathbf{x}_{1}$,  $\dots$, $\mathbf{x}_{T+1}$,
where $\mathbf{x}_{0} = [\mathbf{v}_W; \mathrm{Atten}(\mathbf{V}, \mathbf{0})]$
is the concatenation of the encoder's last hidden state $\mathbf{v}_W$
and the attention output with guidance equals to zero; and
$\mathbf{x}_{i} = [\psi(s_{i-1}); \mathrm{Atten}(\mathbf{V}, \mathbf{h}'_{i-1})]$, for $i=1,\dots,T+1$,
is made up of the embedding $\psi()$ of the $(i-1)$-th token $s_{i-1}$
and the attention output guided by the hidden state of decoder LSTMs in the previous time-step $\mathbf{h}'_{i-1}$.
The embedding function $\psi()$ is defined as a linear layer followed by a $\mathrm{tanh}$
non-linearity.

The attention function $\mathbf{c}_{i} = \mathrm{Atten}(\mathbf{V},\mathbf{h}'_{i})$
is defined as follows:
\begin{equation}
\begin{cases}
\mathbf{g}_j = \tanh( \mathbf{W}_v \mathbf{v}_j+ \mathbf{W}_h \mathbf{h}'_{i}), \,\, j= 1,\dots,W, \\
\boldsymbol{\alpha} = \mathrm{softmax}(\mathbf{w}_g^\T \cdot [\mathbf{g}_1, \mathbf{g}_2, \dots, \mathbf{g}_W]), \\
\mathbf{c}_{i} = \sum_{j=1}^{W} \alpha_j \mathbf{v}_j,
\end{cases}
\end{equation}
where $\mathbf{V} = [\mathbf{v}_1, \dots, \mathbf{v}_W]$ is the variable-length sequence of features to be attended, $\mathbf{h}'_{i}$ is the guidance vector, $\mathbf{W}_v$ and $\mathbf{W}_h$
are linear embedding weights to be learned, $\boldsymbol{\alpha}$ is the attention weights of size $W$, and $\mathbf{c}_{i}$ is a weighted sum of input features.

At each time-step $t = 0, 1, \dots, T+1$, the decoder LSTMs compute their hidden state $\mathbf{h}'_t$ and
output vector $\mathbf{y}_t$ as follows:
\begin{equation}
\begin{cases}
\mathbf{h}'_t = \mathrm{f}(\mathbf{x}_t, \mathbf{h}'_{t-1}), \\
\mathbf{y}_t = \mathrm{\varphi(\mathbf{h}'_t)} = \mathrm{softmax}(\mathbf{W}_o \mathbf{h}'_t)
\end{cases}
\end{equation}
where the LSTM~\cite{LSTM} is used for the recurrence formula $\mathrm{f}()$, and
$\mathbf{W}_o$ linearly transforms hidden states to the output space of size $38$,
including $26$ case-insensitive characters, $10$ digits, a token representing all punctuations like ``!'' and ``?'', and a special END token.

At test time, the token with the highest probability in previous output $\mathbf{y}_{t}$ is selected as the input token at step $t+1$, instead of the ground-truth tokens $s_1, \dots, s_T$.
The process is started with the START token, and repeated until we get the special END token.

\subsection{Loss Functions and Training}

\noindent {\bf Loss Functions}
As we demonstrate above, our system takes as input an image, word bounding boxes and their labels during training.
Both TPN and TDN employ the binary logistic loss $L_{{cls}}$ for classification, and smooth $L_1$ loss $L_{{reg}}$~\cite{renNIPS15fasterrcnn} for regression.
So the loss for training TPN is
\begin{equation}\label{eq1}
L_{{T\!P\!N}}=\frac{1}{N} \sum_{i=1}^{N} {L}_{{cls}} (p_i,p_i^\star)  + \frac{1}{N_{+}}
\sum_{i=1}^{N_+} L_{{reg}} (\mathbf{d}_i,\mathbf{d}_i^\star),
\end{equation}
where $N$ is the number of randomly sampled anchors in a mini-batch and $N_+$ is the number of positive anchors in this batch
(the range of positive anchor indices is from $1$ to $N_+$).
The mini-batch sampling and training process of TPN are similar to that used in~\cite{renNIPS15fasterrcnn}.
An anchor is considered as positive if its Intersection-over-Union (IoU) ratio with a ground-truth
is greater than $0.7$ and considered as negative if its IoU with any ground-truth is
smaller than $0.3$.
In this paper, $N$ is set to $256$ and $N_+$ is at most $128$.
$p_i$ denotes the predicted probability of anchor $i$ being text and $p_i^\star$
is the corresponding ground-truth label ($1$ for text, $0$ for non-text).
$\mathbf{d}_i$ is the predicted coordinate offsets $(\mathrm{dx}_i, \mathrm{dy}_i, \mathrm{dw}_i, \mathrm{dh}_i)$ for anchor $i$,
and $\mathbf{d}_i^\star$ is the associated offsets for anchor $i$ relative to the ground-truth.
Bounding box regression is only for positive anchors, as there is no ground-truth bounding box matched with negative ones.

For the final outputs of the whole system, we apply a multi-task loss for both detection and recognition:
\begin{align}\label{eq2}
L_{D\!R\!N} &= \frac{1}{\hat{N}} \sum_{i=1}^{\hat{N}} L_{cls} (\hat{p}_i,\hat{p}_i^\star)
+ \frac{1}{\hat{N}_{+}} \sum_{i=1}^{\hat{N}_+} L_{reg} (\hat{\mathbf{d}}_i,\hat{\mathbf{d}}_i^\star)  \notag \\
&+ \frac{1}{\hat{N}_{+}} \sum_{i=1}^{\hat{N}_{+}} L_{rec} (\mathbf{Y}^{(i)}, \mathbf{s}^{(i)})
\end{align}
where $\hat{N} = 128$ is the number of text proposals sampled from the output of TPN,
and $\hat{N}_{+} \leq 64$ is the number of positive ones.
The thresholds for positive and negative anchors are set to $0.6$ and $0.4$ respectively, which are less strict than those used for training TPN.
In order to mine hard negatives, we first apply TDN on $1000$ randomly sampled
negatives and select those with higher textness scores.
$\hat{p}_i$ and $\hat{\mathbf{d}}_i$ are the outputs of TDN.
$\mathbf{s}^{(i)}$ is the ground-truth tokens for sample $i$ and $\mathbf{Y}^{(i)} = \{ \mathbf{y}^{(i)}_0, \mathbf{y}^{(i)}_1, \dots, \mathbf{y}^{(i)}_{T+1} \}$
is the corresponding output sequence of decoder LSTMs.
$L_{rec}(\mathbf{Y}, \mathbf{s}) = - \sum_{t=1}^{T+1} \log \mathbf{y}_t(s_{t})$ denotes the cross entropy loss on $\mathbf{y}_1, \dots, \mathbf{y}_{T+1}$,
where $\mathbf{y}_t(s_{t})$ represents the predicted probability of the output being $s_t$ at
time-step $t$ and the loss on $\mathbf{y}_0$ is ignored.

Following~\cite{renNIPS15fasterrcnn}, we use an approximate joint training process
to minimize the above two losses together (ADAM~\cite{adam14} is adopted),
ignoring the derivatives with respect to the proposed boxes' coordinates.

\noindent{\bf Data Augmentation} We sample one image per iteration in the training phase. Training images are resized to shorter side of $600$ pixels and longer side of at most $1200$ pixels. Data augmentation is also implemented to improve the robustness of our model, which includes:

$1)$ randomly rescaling the width of the image by ratio $1$ or $0.8$ without changing its height, so that the bounding boxes have more variable aspect ratios;

$2)$ randomly cropping a subimage which includes all text in the original image,  padding with $100$ pixels on each side, and resizing to $600$ pixels on shorter side.

\noindent{\bf Curriculum Learning}
In order to improve generalization and accelerate the convergence speed,
we design a curriculum learning~\cite{Bengio2009} paradigm to train the model from gradually more complex data.

$1)$ We generate $48$k images containing words in the ``Generic'' lexicon~\cite{Max2016IJCV} of size $90$k by using the synthetic engine proposed in~\cite{Gupta16}.
The words are randomly placed on simple {\em pure colour backgrounds} ($10$ words per image on average). We lock TRN initially, and train the rest parts of our proposed model on these synthetic images in the first $30$k iterations,
with convolutional layers initialized from the trained VGG-$16$ model and other parameters
randomly initialized according to Gaussian distribution.
For efficiency, the first four convolutional layers are fixed during the entire training process.
The learning rate is set to $10^{-5}$ for parameters in the rest of convolutional layers and $10^{-3}$ for randomly initialized parameters.

$2)$ In the next $30$k iterations,
TRN is added and trained with a learning rate of $10^{-3}$, together with other parts in which the learning rate for randomly initialized parameters is halved to $5 \times 10^{-4}$. We still use the $48$k synthetic images as they contain a comprehensive $90$k word vocabulary. With this synthetic dataset, a character-level language model can be learned by TRN.

$3)$ In the next $50$k iterations,
the training examples are randomly selected from
the ``Synth800k"~\cite{Gupta16} dataset,
which consists of $800$k images with averagely $10$ synthetic words placed on each {\em real scene background}.
The learning rate for convolutional layers remains at $10^{-5}$, but that for others is halved to
$10^{-4}$.

$4)$ Totally $2044$ {\em real-world} training images from ICDAR2015~\cite{icdar2015}, SVT~\cite{Wangkai2011} and AddF2k~\cite{Zhong2016} datasets
are employed for another $20$k iterations. In this stage, all the convolutional layers are fixed and the learning rate for others is further halved to $10^{-5}$.
These real images contain much less words than synthetic ones, but their
appearance patterns are much more complex.

\section{Experiments}
\label{SEC:Exp}
In this section, we perform experiments to verify the effectiveness of the proposed method.All experiments are implemented on an NVIDIA Tesla M$40$ GPU with $24$GB memory.
We rescale the input image into multiple sizes during test phase in order to cover the large range of bounding box scales, and sample $300$ proposals with the highest textness scores produced by TPN.
The detected bounding boxes are then merged via NMS according to their textness scores and fed into TRN for recognition.

\noindent {\bf Criteria} We follow the evaluation protocols used in ICDAR$2015$ Robust Reading Competition~\cite{icdar2015}: a bounding box is considered as correct if its
IoU ratio with any ground-truth is greater than $0.5$ and the recognized word also matches, ignoring the case.
The words that contain alphanumeric characters and
no longer than three characters are ignored.
There are two evaluation protocols used in the task of scene text spotting: ``End-to-End" and ``Word Spotting". ``End-to-End" protocol requires that all words in the image are to be recognized, with independence of whether the string exists or not in the provided contextualised lexicon, while ``Word Spotting" on the other hand, only looks at the words that actually exist in the lexicon provided, ignoring all the rest that do not appear in the lexicon. %

\noindent {\bf Datasets}
The commonly used datasets for scene text spotting include ICDAR$2015$~\cite{icdar2015}, ICDAR$2011$~\cite{icdar2011} and Street View Text (SVT)~\cite{Wangkai2011}.
We use the dataset for the task of ``Focused Scene Text" in ICDAR$2015$ Robust Reading Competition, which consists of $229$ images for training and $233$ images for test.
In addition, it provides $3$ specific lists of words as lexicons for reference in the test phase, \ie, ``Strong'', ``Weak'' and ``Generic''. ``Strong'' lexicon provides $100$ words per-image including all words appeared in the image. ``Weak'' lexicon contains all words appeared in the entire dataset, and ``Generic'' lexicon is a $90$k word vocabulary proposed by~\cite{Max2016IJCV}. %
ICDAR$2011$ does not provide any lexicon. So we only use the $90$k vocabulary as context.
SVT dataset consists of $100$ images for training and $249$ images for test. These images are harvested from Google Street View and often have a low resolution. %
It also provides a ``Strong'' lexicon with $50$ words per-image. %
As there are unlabelled words in SVT, we only evaluate the ``Word-Spotting'' performance on this dataset.

\subsection{Evaluation under Different Model Settings}

\begin{table*}[h]
	\
	\newcommand{\tabincell}[2]{\begin{tabular}{@{}#1@{}}#2\end{tabular}}
	\begin{center}
		\caption{Text spotting results on different benchmarks. We present the F-measure here in percentage. ``Ours Two-stage'' uses separate models for detection and recognition, while other ``Ours'' models are end-to-end trained.  %
		``Ours Atten+Vary'' achieves the best performance on almost all datasets. }
		\label{Tab:7}
		\scalebox{0.9}{
			\begin{tabular}{l|c|c|c|c|c|c|c|c|cc}
				\hline
				Method & \multicolumn{3}{|c}{\tabincell{c}{ICDAR$2015$ \\ Word-Spotting}} &  \multicolumn{3}{|c} {\tabincell{c}{ICDAR$2015$ \\ End-to-End}} & \multicolumn{1}{|c}{\tabincell{c}{ICDAR$2011$ \\ Word-Spotting}} & \multicolumn{2}{|c}{\tabincell{c}{SVT \\ Word-Spotting}} & \\\cline{2-10} 	& \multicolumn{1}{c|}{Strong} & \multicolumn{1}{|c|}{Weak}  & \multicolumn{1}{|c|}{Generic}   & \multicolumn{1}{c|}{Strong} & \multicolumn{1}{|c|}{Weak} & \multicolumn{1}{|c|}{Generic} &  \multicolumn{1}{c|}{Generic} & \multicolumn{1}{|c|}{Strong}  &  \multicolumn{1}{|c}{Generic} \\
				\hline
				Deep2Text II+~\cite{Yin2014pami} & $84.84$ & $83.43$ & $78.90$ & $81.81$ & $79.47$ & $76.99$ & $-$ & $-$  & $-$\\
				\hline
				Jaderberg~\etal~\cite{Max2016IJCV} & $90.49$ & $-$ & $76$ & $86.35$ & $-$ & $-$  & $76$ & $76$ & $53$ \\
				\hline
				FCRNall+multi-filt~\cite{Gupta16} & $-$ & $-$ & $84.7$ & $-$ & $-$ & $-$ & $84.3$ & $67.7$ & $55.7$  \\
				\hline
				TextBoxes~\cite{LiaoAAAi2017} & $93.90$ & $91.95$ & $85.92$ & $\textbf{91.57}$ & $89.65$ & $83.89$ & $87$ & $84$ & $64$ \\
				\hline
				YunosRobot$1.0$ & $86.78$ & $-$ & $86.78$ & $84.20$ & $-$ & $84.20$  & $-$ & $-$  & $-$\\
				\hline
				\hline
				\tabincell{c}{Ours Two-stage} & $92.94 $ & $90.54 $ & $84.24$  & $88.20$ & $86.06$ & $81.97$ & $82.86$ & $82.19$ & $62.35$  \\
				\hline
				\tabincell{c}{Ours NoAtten+Fixed} & $92.70$ & $90.37$ & $83.83$  & $87.73$ & $85.53$ & $79.18$ & $81.70$ & $79.49$ & $58.70$  \\
				\hline
				\tabincell{c}{Ours Atten+Fixed} & $93.33 $ & $91.66 $ & $87.73$  & $90.72 $ & $87.86$ & $83.98 $ & $83.81$ & $81.80$ & $64.50$  \\
				\hline
				\tabincell{c}{Ours Atten+Vary} & $\textbf{94.16}$ & $\textbf{92.42}$ & $\textbf{88.20}$  & $91.08$ & $\textbf{89.81}$ & $\textbf{84.59}$ & $\textbf{87.70}$ & $\textbf{84.91}$ & $\textbf{66.18}$  \\
				\hline
			\end{tabular}
		}
	\end{center}
\vspace{-0.5cm}
\end{table*}

In order to show the effectiveness of our proposed varying-size RoI pooling (see Section \ref{sec:rfe})
and the attention mechanism (see Section \ref{sec:trn}),
we examine the performance of our model with different settings in this subsection.
With the fixed RoI pooling size of $4 \times 20$,
we denote the models with and without the attention mechanism
as ``Ours Atten+Fixed'' and ``Ours NoAtten+Fixed'' respectively.
The model with both attention and varying-size RoI pooling is denoted as ``Ours Atten+Vary",
in which the size of feature maps after pooling is calculated by Equ.~\eqref{eq:pool}
with $H = 4$ and $W_{max} = 35$.

\begin{figure}
	\begin{center}
		\includegraphics[width=0.48\textwidth]{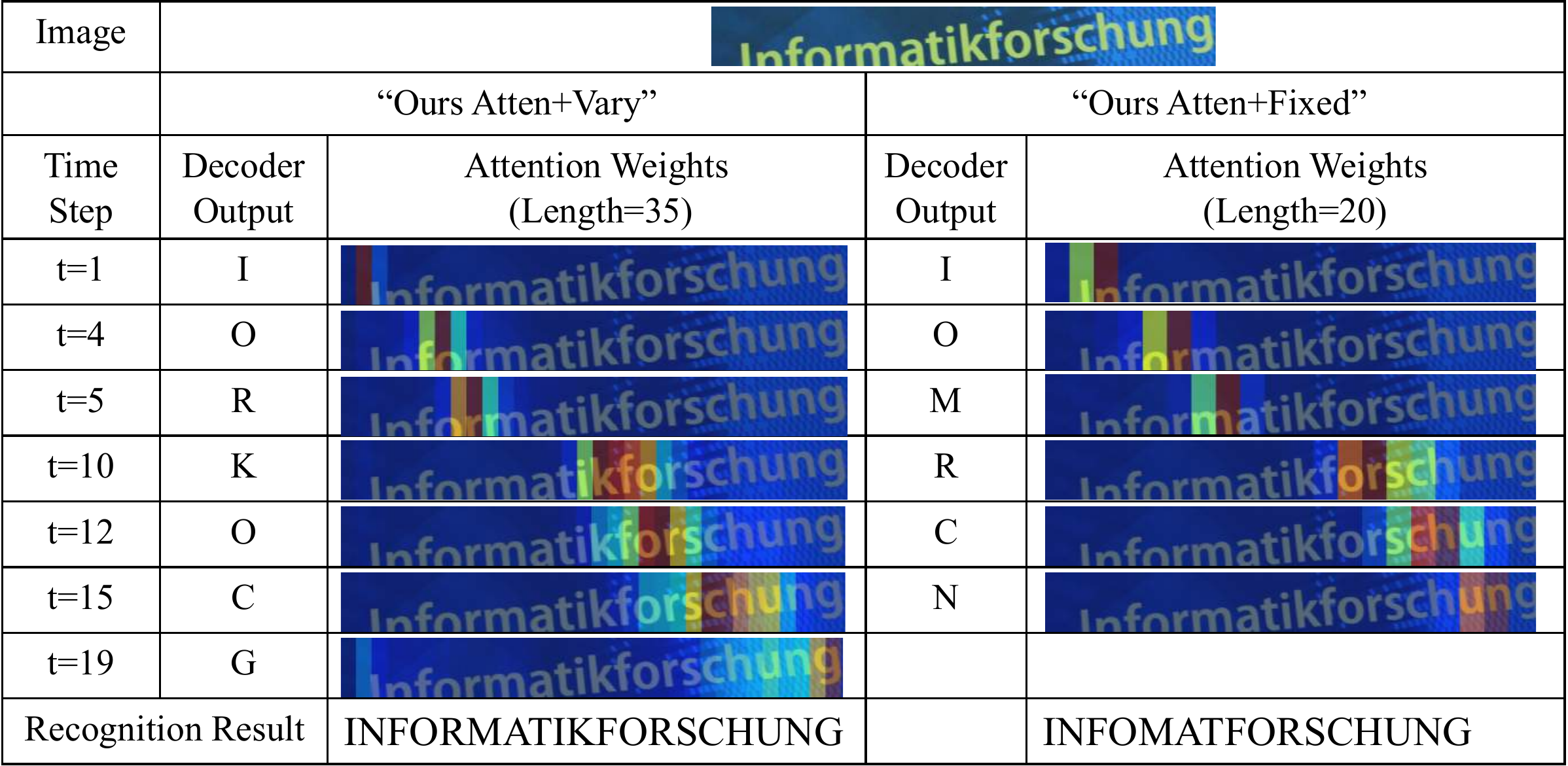}
	\end{center}
\vspace{-0.3cm}
	\caption{Attention mechanism based sequence decoding process by ``Ours Atten+Vary'' and ``Ours Atten+Fixed'' separately. The heat maps show that at each time step, the position of the character to be decoded has higher attention weights, so that the corresponding local features will be extracted and assist the text recognition. However, if we use the fixed size RoI pooling, information may be lost during pooling, especially for a long word, which leads to an incorrect recognition result. In contrast, ``Ours Atten+Vary'' gives the correct result, even if some parts of the word image are missed, such as ``I'', ``n'' in this example.}
	\label{fig:attention}
	\vspace{-0.5cm}
\end{figure}

Although the last hidden state of LSTMs encodes the holistic information of RoI image patch, it still lacks details. Particularly for a long word image patch, the initial information may be lost during the recurrent encoding process.
Thus, we keep the hidden states of encoder LSTMs at each time step as context. The  attention model can choose the corresponding local features for each character during decoding process, as illustrated in Figure~\ref{fig:attention}.
From Table~\ref{Tab:7}, we can see that the model with attention mechanism,
namely ``Ours Atten+Fixed'', achieves higher F-measures on all evaluated data than ``Ours NoAtten+Fixed'' which does not use attention.

One contribution of this work is a new region feature encoder,
which is composed of a varying-size RoI pooling mechanism and an LSTM sequence encoder.
To validate its effectiveness, we compare the performance of models ``Ours Atten+Vary"
and ``Ours Atten+Fixed".
Experiments shows that varying-size RoI pooling performs significantly better for long words.
For example, ``Informatikforschung" can be recognized correctly by ``Ours Atten+Vary", but not by ``Ours Atten+Fixed" (as shown in Figure~\ref{fig:attention}), because
a large portion of information for long words is lost by fixed-size RoI pooling.
As illustrated in Table~\ref{Tab:7}, adopting varying-size RoI pooling (``Ours Atten+Vary")
instead of fixed-size pooling (``Ours Atten+Fixed")
makes F-measures increase around $1\%$ for ICDAR$2015$, $4\%$ for ICDAR$2011$ and $3\%$ for SVT with strong lexicon used.

\subsection{Joint Training vs. Separate Training}

Previous works~\cite{Max2016IJCV, Gupta16, LiaoAAAi2017} on text spotting typically
perform in a two-stage manner,
where detection and recognition are trained and processed separately.
The text bounding boxes detected by a model need to be cropped from the image and then recognized by another model.
In contrast, our proposed model is trained jointly for both detection and recognition.
By sharing convolutional features and RoI encoder,
the knowledge learned from the correlated detection and recognition tasks can be transferred between each other and results in better performance for both tasks.

To compare with the model ``Ours Atten+Vary" which is jointly trained,
we build a two-stage system (denoted as ``Ours Two-stage'') in which detection and recognition models are trained separately.
For fair comparison, the detector in ``Ours Two-stage'' is built by
removing the recognition part from model ``Ours Atten+Vary'' and trained only with the detection objective (denoted as ``Ours DetOnly'').
As to recognition, we employ CRNN~\cite{ShiBY15} that produces state-of-the-art performance on text recognition.
Model ``Ours Two-stage'' firstly adopts ``Ours DetOnly'' to detect text with the same multi-scale inputs. CRNN is then followed to recognize the detected bounding boxes.
We can see from Table~\ref{Tab:7} that model ``Ours Two-stage'' performs worse than ``Ours Atten+Vary"
on all the evaluated datasets.

\begin{figure*}[th]
	\vspace{-0.0cm}
	\centering
	\begin{minipage}[t]{0.7\textwidth}
		\centering
		\includegraphics[width=1\textwidth]{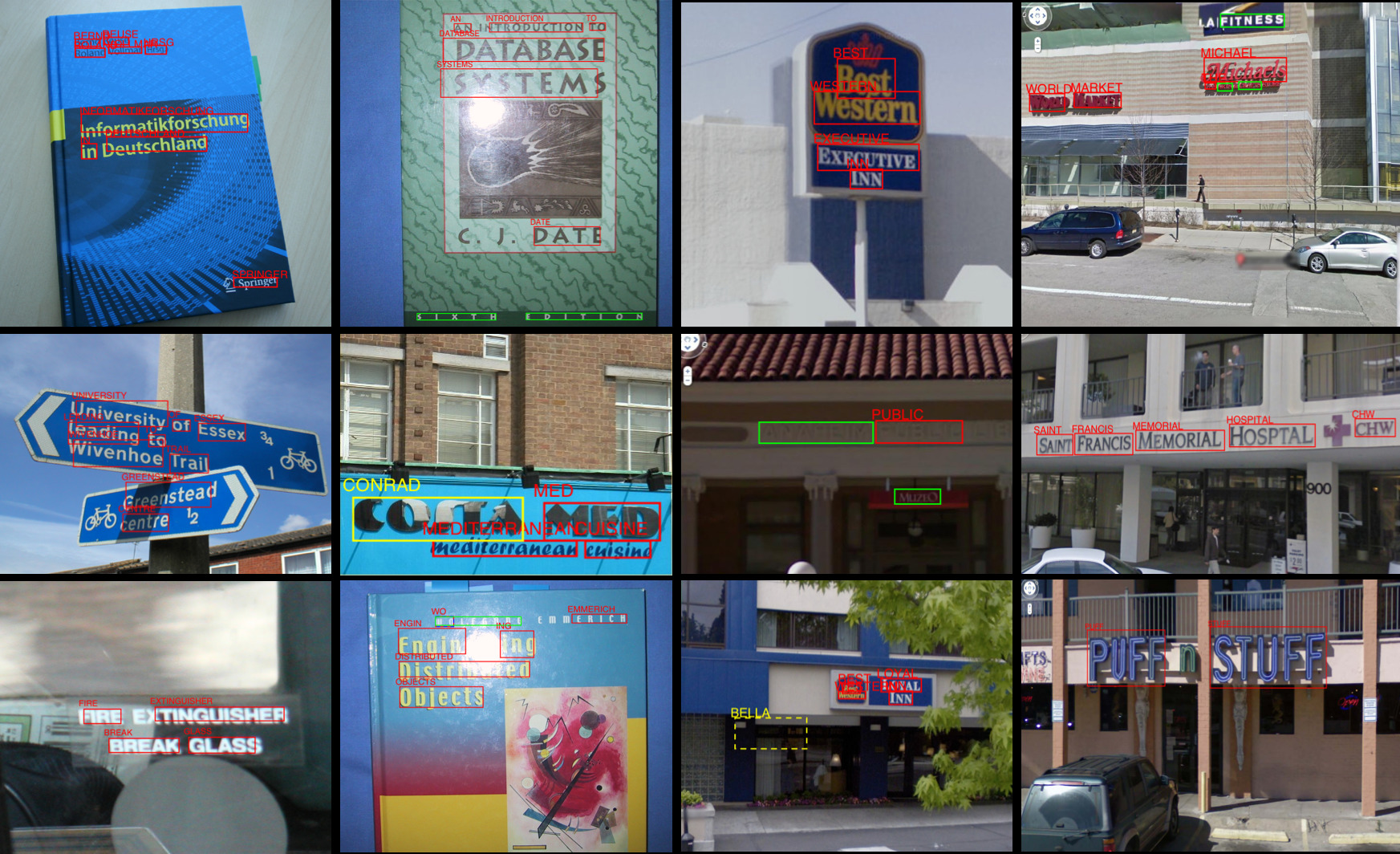}
	\end{minipage}\hfill
	\begin{minipage}[t]{0.28\textwidth}
		\centering
		\vspace{-7.4cm}
		\caption{Examples of text spotting results by ``Ours Atten+Vary''. The first two columns are images from ICDAR$2015$, and the rest are images from SVT. The red bounding boxes are both detected and recognized correctly. The green bounding boxes are missed words, and the yellow dashed bounding boxes are false positives. Note that our recognition network can produce a correct result even if the detected bounding box misses some parts of the word, such as ``INTRODUCTION'' in the $2$nd image of the first row, ``EXTINGUISHER'' in the $1$st image of the last row. That is because our recognition framework learns a language model from the synthetic data. It can infer the word even if some parts are not covered by the bounding box. }
		\label{fig:examples}
	\end{minipage}
\end{figure*}

Furthermore, we also compare the detection-only performance of these two systems.
Note that ``Ours DetOnly'' and the detection part of ``Ours Atten+Vary"
share the same architecture, but they are trained with different strategies:
``Ours DetOnly'' is optimized with only the detection loss, while
``Ours Atten+Vary" is trained with a multi-task loss for both detection and recognition.
In consistent with the ``End-to-End'' evaluation criterion, a detected bounding box is considered to be correct if its IoU ratio with any ground-truth is greater than $0.5$.
The detection results are presented in Table~\ref{Tab:2}. Without any lexicon used, ``Ours Atten+Vary'' produces a detection performance with F-measures of $85.6\%$ on ICDAR$2015$ and $85.1\%$ on ICDAR$2011$, which are averagely $2\%$ higher than those given by ``Ours DetOnly''. This result illustrates that detector performance can be improved
via joint training.

\begin{table}
	\
	\newcommand{\tabincell}[2]{\begin{tabular}{@{}#1@{}}#2\end{tabular}}
	\begin{center}
		\caption{Text detection results on different datasets. Precision (P) and Recall (R) at maximum F-measure (F) are reported in percentage.
		The jointly trained model (``Ours Atten+Vary") gives better detection results than the one trained with detection loss only (``Ours DetOnly").
			}
		\label{Tab:2}
		{
			\scalebox{0.82}{
				\begin{tabular}{l|c c c|c c c}
					\hline
					\multirow{2}{*}{Method} & \multicolumn{3}{|c}{ICDAR$2015$} &   \multicolumn{3}{|c}{ICDAR$2011$}   \\ \cline{2-7}	& \multicolumn{1}{c}{R} & \multicolumn{1}{c}{P}  & \multicolumn{1}{c|}{F}   & \multicolumn{1}{c}{R} & \multicolumn{1}{c}{P}  & \multicolumn{1}{c}{F}
					\\
					\hline
					Jaderberg~\etal~\cite{Max2016IJCV} & $68.0$  & $86.7$ & $76.2$ & $69.2$ & $87.5$ & $77.2$  \\
					\hline
					FCRNall+multi-filt~\cite{Gupta16} & $76.4$  & $\textbf{93.8}$ & $84.2$ & $76.9$ & $\textbf{94.3}$ & $84.7$  \\
					\hline
					\tabincell{c}{Ours DetOnly}  & $78.5$  & $88.9$ & $83.4$ & $80.0$ & $87.5$ & $83.5$   \\
					\hline
					\tabincell{c}{Ours Atten+Vary}  & $\textbf{80.5}$  & $91.4$ & $\textbf{85.6}$  & $\textbf{81.7}$ & $89.2$ & $\textbf{85.1}$  \\
					\hline
				\end{tabular}
			}
		}

	\end{center}
\vspace{-0.5cm}
\end{table}

\subsection{Comparison with Other Methods}

In this part, we compare the text spotting results of ``Ours Atten+Vary'' with other state-of-the-art approaches. As shown in Table~\ref{Tab:7}, ``Ours Atten+Vary'' outperforms all compared methods on most of the evaluated datasets. In particular, our method shows an significant superiority when using a generic lexicon. It leads to a $1.5\%$ higher recall on average than the state-of-the-art TextBoxes~\cite{LiaoAAAi2017}, using only $3$ input scales compared with $5$ scales used by TextBoxes.

Several text spotting examples are presented in Figure~\ref{fig:examples}, which demonstrate that model ``Ours Atten+Vary'' is capable of dealing with words of different aspect ratios and orientations.
In addition, our system is able to recognize words even if their bounding boxes do not cover the whole words, as it potentially learned a character-level language model from the synthetic data.

\subsection{Speed}
Using an M40 GPU, model ``Ours Atten+Vary'' takes approximately $0.9s$ to process
an input image of $600 \times 800$ pixels.
It takes nearly $0.45s$ to compute the convolutional features, $0.02s$ for text proposal calculation, $0.25s$ for RoI encoding, $0.01s$ for text detection and $0.15s$ for word recognition. On the other hand, model ``Ours Two-stage'' spends around $0.45s$ for word recognition on the same detected bounding boxes, as it needs to crop the word patches, and re-calculate the convolutional features during recognition.

\section{Conclusion}
\label{SEC:Con}

In this paper we presented a unified end-to-end trainable DNN for simultaneous text detection and recognition in natural scene images. A novel RoI encoding method was proposed, considering the large diversity of aspect ratios of word bounding boxes. With this framework,  scene text  can be detected and recognized in a single forward pass efficiently and accurately.
Experimental results illustrate that the proposed method can produce impressive performance on standard benchmarks.
One of potential future works is on handling images with multi-oriented text.

\section{Appendix}

\subsection{Training Data with Different Levels of Complexity}
\label{sec:Learning}

In this paper, we design a curriculum learning~\cite{Bengio2009} paradigm to train the model from gradually more complex data. Here, we would like to give a detailed introduction to the used training data.

\begin{figure}[h]
	\begin{center}
		\includegraphics[width=0.45\textwidth]{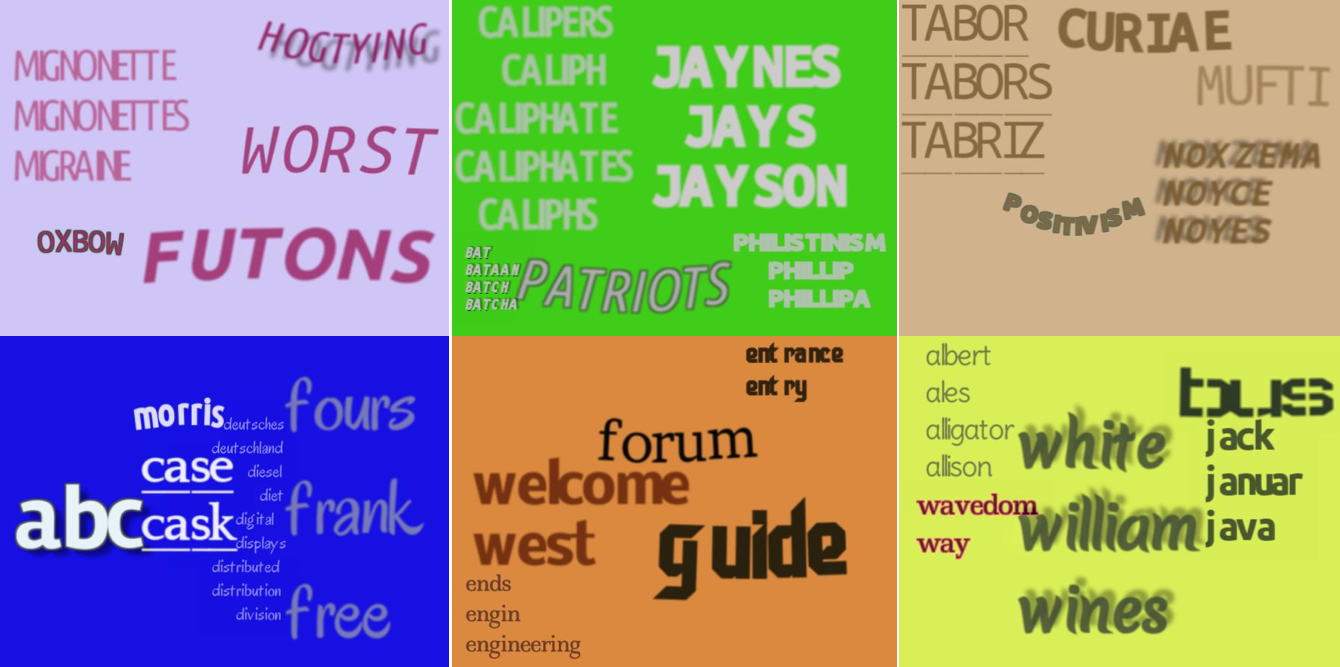}
	\end{center}
	\caption{Synthetic images with words randomly placed on simple pure colour backgrounds. These images are used to pre-train our model. %
	}
	\label{fig:painimg}
\end{figure}

Firstly,  we generate $48$k images containing words in the ``Generic'' lexicon~\cite{Max2016IJCV} of size $90$k by using the synthetic engine proposed in~\cite{Gupta16}. The words are randomly placed on simple {\em pure colour backgrounds}, as shown in Figure~\ref{fig:painimg}.
Note that these $48$k images contain a comprehensive word vocabulary, so a character-level language model can be learned by Text Recognition Network (TRN).

\begin{figure}[h]
	\begin{center}
		\includegraphics[width=0.45\textwidth]{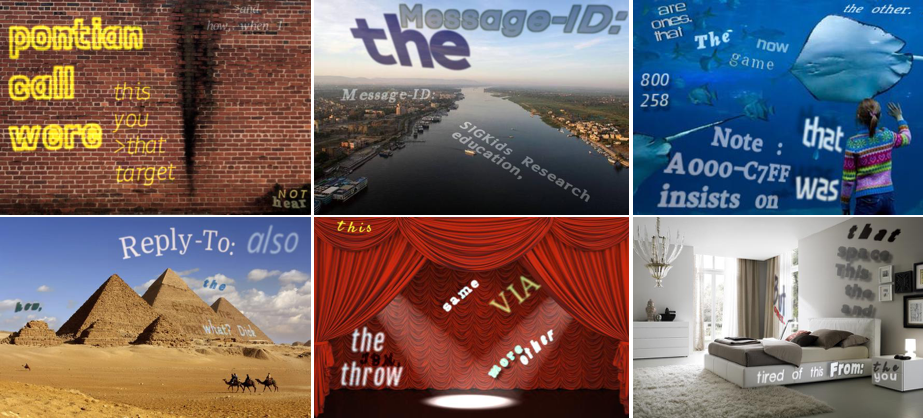}
	\end{center}
	\caption{Synthetic images from~\cite{Gupta16}, which are created via blending rendered words into real natural scene images. The background is more complex
		than that of images in Figure~\ref{fig:painimg}. %
	}
	\label{fig:synimg}
\end{figure}

Next,
the ``Synth800k"~\cite{Gupta16} dataset is used to further tune the model,
which contains $800$k images created via blending rendered words into real natural scenes, as presented in  Figure~\ref{fig:synimg}.
These images have more complex background, so that the model will be further fine-tuned to handle complicated appearance patterns.

\begin{figure}[h]
	\begin{center}
		\includegraphics[width=0.45\textwidth]{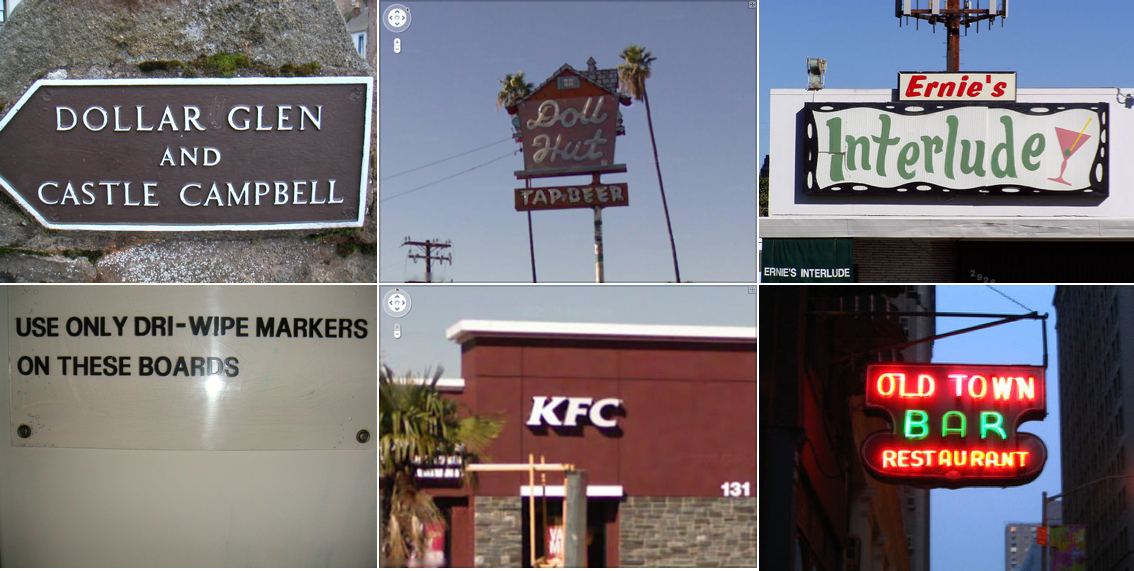}
	\end{center}
	\caption{The real-world images captured from natural scene,
		     with the $1$st, $2$nd and $3$rd columns corresponding to
		     images from datasets ICDAR2015, SVT and AddF2k respectively.
	}
	\label{fig:realimg}
\end{figure}

Finally, we use  $2044$ {\em real-world} images to fine-tune our model. %
They are naturally captured images explicitly focusing around the text of interest, as shown in Figure~\ref{fig:realimg}. %

\subsection{Varying-size Pooling vs. Fixed-size Pooling}
\label{sec:pooling}

\begin{figure}[h]
	\begin{center}
		\subfigure[]{
		\includegraphics[width=0.45\textwidth]{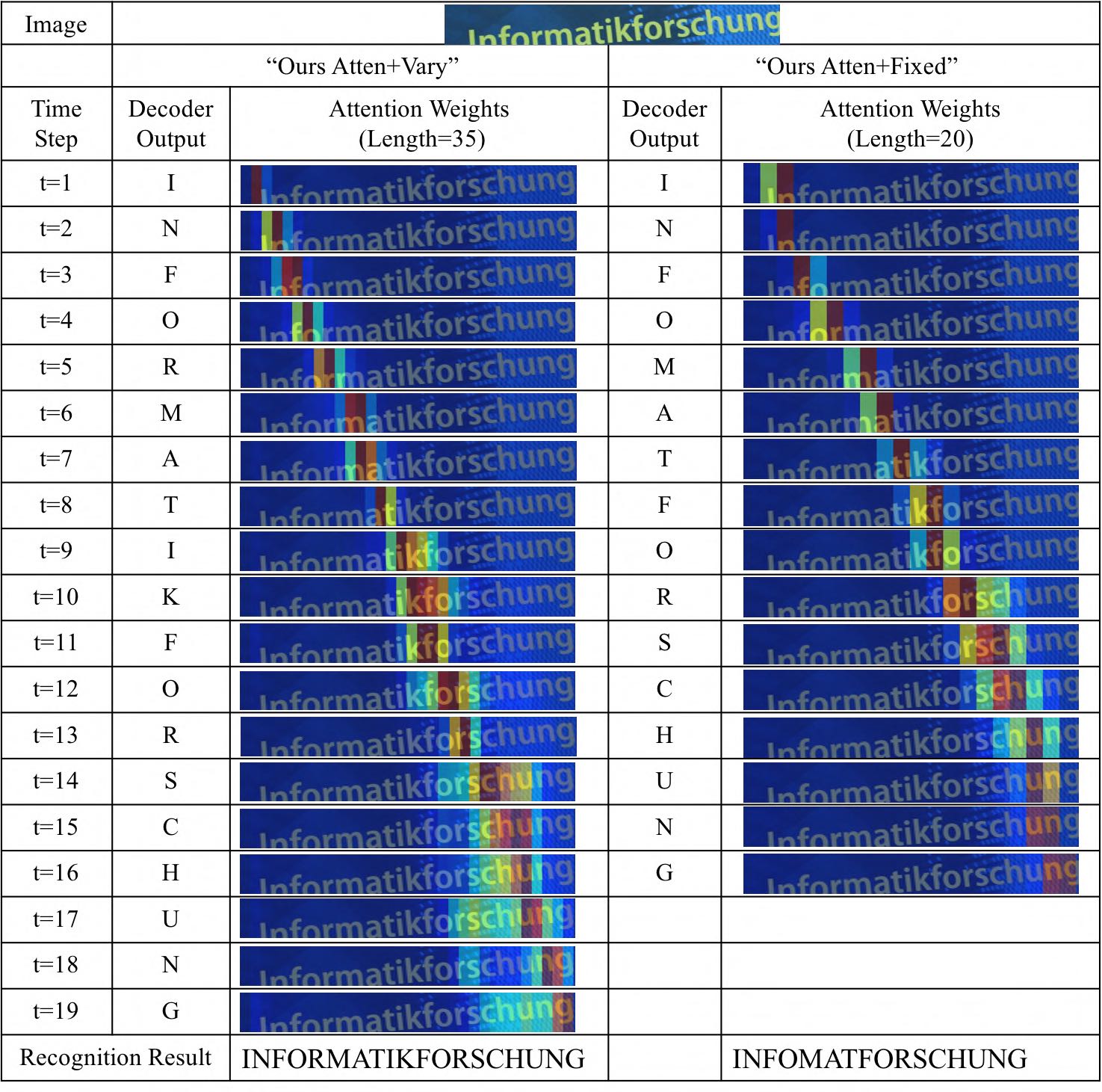}
		}
		\subfigure[]{
		\includegraphics[width=0.45\textwidth]{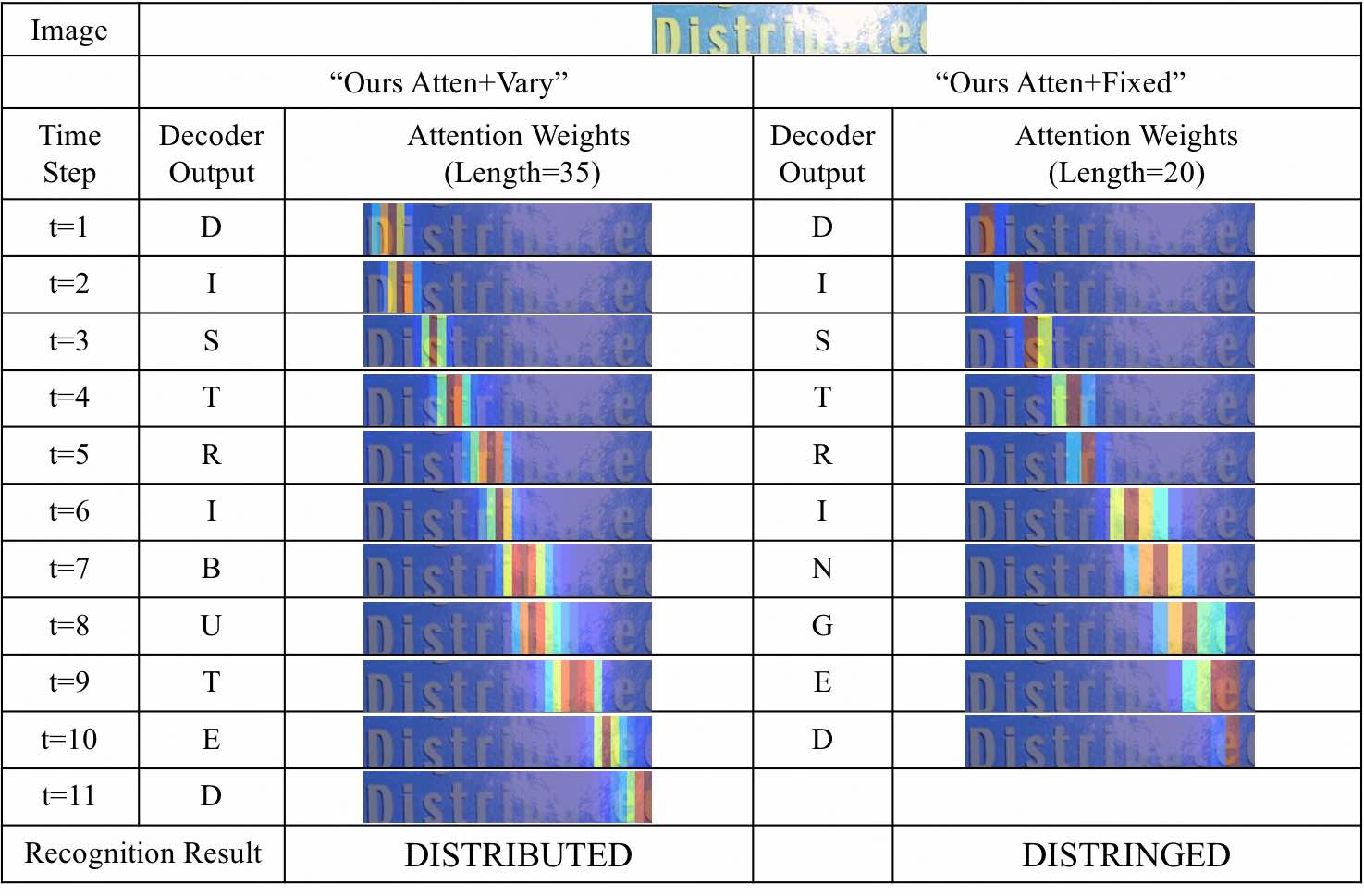}
	    }
	\end{center}
	\caption{
		Attention based decoding process of ``Ours Atten+Vary'' and ``Ours Atten+Fixed''. The heat maps demonstrate the attention weights at each time step of RNN encoding.}
	\label{fig:att1}
\end{figure}

In this work, we propose a new method for encoding an image region
of variable size into a fixed-length representation.
Unlike the conventional method~\cite{RGB2015ICCV, renNIPS15fasterrcnn}, where each Region-of-Interest (RoI) is pooled into a fixed-size feature map, we pool RoIs according to their aspect ratios. Here we present more experimental results in
Figure~\ref{fig:att1} to verify the effectiveness of our proposed encoding method.
Compared to fixed-size pooling,
our method (varying-size pooling) divide image regions for long words into more
parts ($35$ versus $20$), such that information for every character is reserved.

\subsection{Additional Text Spotting Results}
\label{sec:res}

\begin{figure*}
	\begin{center}
		\subfigure[Results on ICDAR$2015$ Images]{
			\includegraphics[width=0.9\textwidth]{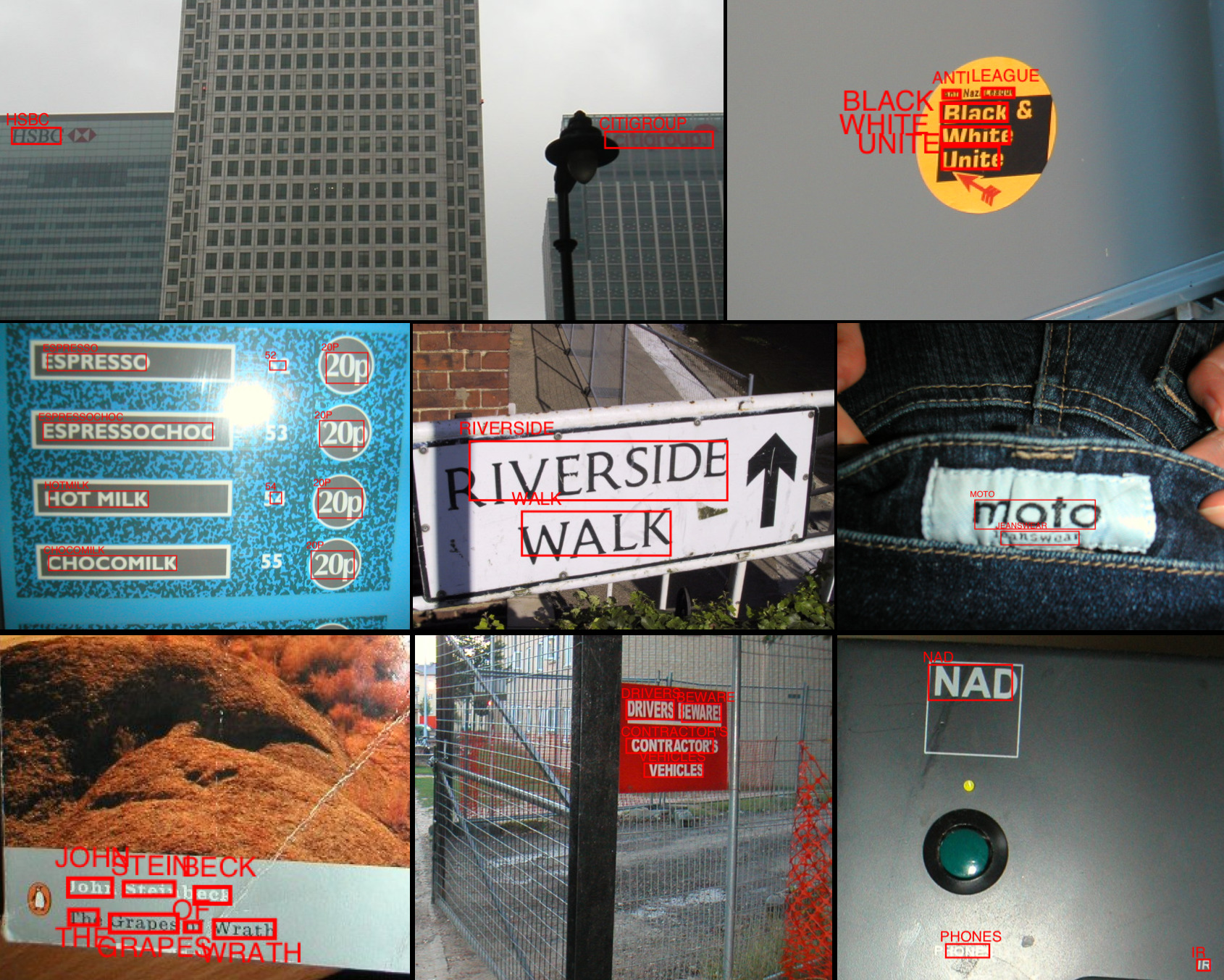}
		}
		\subfigure[Results on SVT Images]{
			\includegraphics[width=0.9\textwidth]{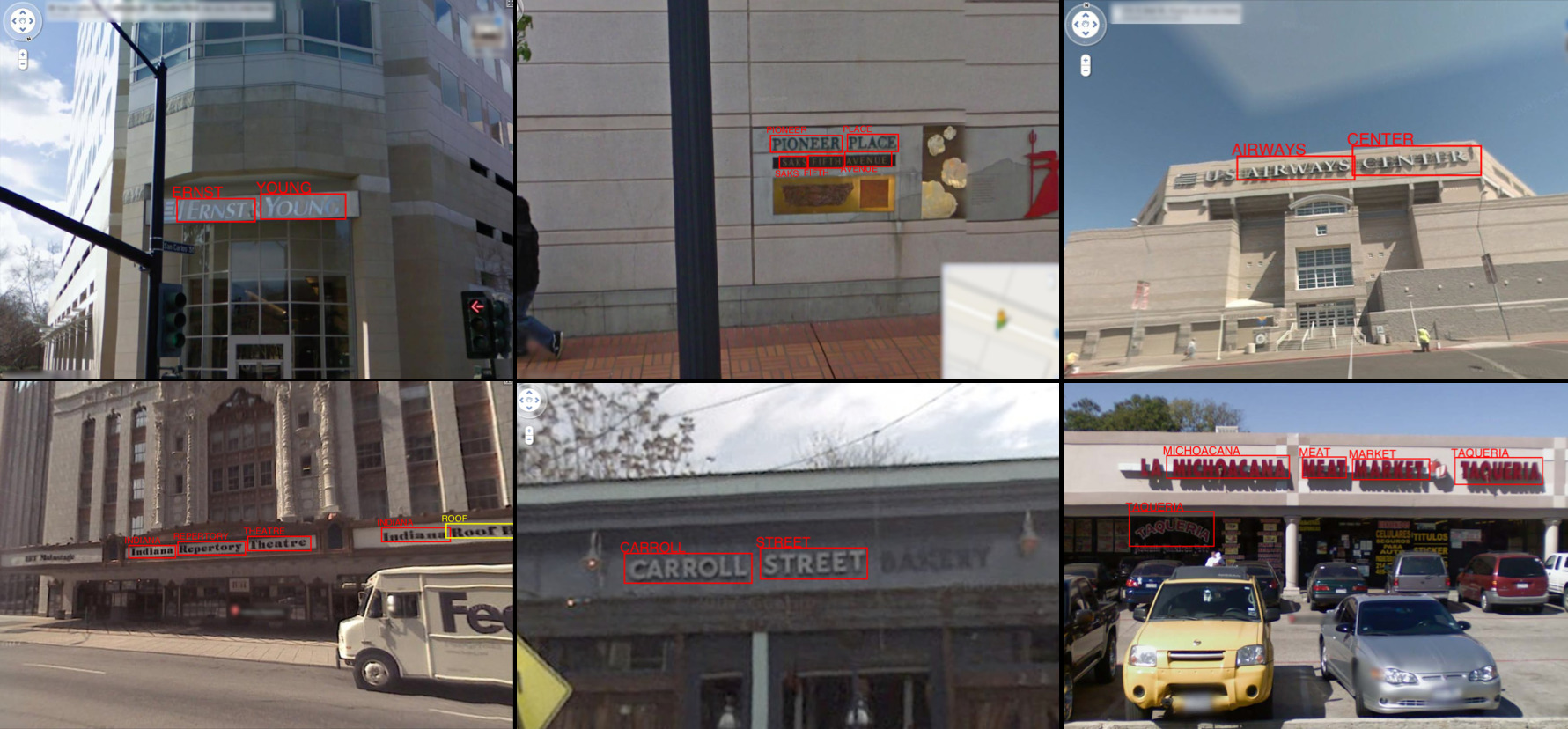}
		}
	\end{center}
	\caption{Text spotting results on different datasets,
	produced by our proposed model ``Ours Atten+Vary''.}
	\label{fig:res}
\end{figure*}

Here we present more test results by our proposed model, as shown in Figure~\ref{fig:res}.

\clearpage

{\small
	\bibliographystyle{ieee}
	\bibliography{mybibfile}
}

\end{document}